\documentclass[11pt,a4paper]{article}
\usepackage{acl2015}

\usepackage{latexsym}
\usepackage{url}
\usepackage{amsmath,amssymb}
\usepackage{ifthen}
\usepackage{graphicx}
\usepackage{stmaryrd}
\usepackage{enumitem}
\usepackage{multicol}
\usepackage{multirow}
\usepackage[usenames,dvipsnames,svgnames,table]{xcolor}
\usepackage{subfigure}
\usepackage{algorithm}
\usepackage{algorithmic}
\usepackage{xspace}

\newcommand\sC{\ensuremath{\mathcal{C}}}

\newcommand\sE{\ensuremath{\mathcal{E}}}

\newcommand\sG{\ensuremath{\mathcal{G}}}

\newcommand\sN{\ensuremath{\mathcal{N}}}

\newcommand\sR{\ensuremath{\mathcal{R}}}

      \newcommand\pc[1]{\ensuremath{\left\{ #1 \right\}}}  

\newcommand\R{\ensuremath{\mathbb{R}}}      \newcommand\eqdef{\ensuremath{\stackrel{\rm def}{=}}}    \newcommand\refeqn[1]{(\ref{eqn:#1})}

\newcommand\refsec[1]{Section~\ref{sec:#1}}

\newcommand\reffig[1]{Figure~\ref{fig:#1}}

\newcommand\reftab[1]{Table~\ref{tab:#1}}

\ifthenelse{\isundefined{\definition}}{\newtheorem{definition}{Definition}}{}
\ifthenelse{\isundefined{\assumption}}{\newtheorem{assumption}{Assumption}}{}
\ifthenelse{\isundefined{\proposition}}{\newtheorem{proposition}{Proposition}}{}
\ifthenelse{\isundefined{\theorem}}{\newtheorem{theorem}{Theorem}}{}
\ifthenelse{\isundefined{\lemma}}{\newtheorem{lemma}{Lemma}}{}
\ifthenelse{\isundefined{\corollary}}{\newtheorem{corollary}{Corollary}}{}
\ifthenelse{\isundefined{\alg}}{\newtheorem{alg}{Algorithm}}{}
\ifthenelse{\isundefined{\example}}{\newtheorem{example}{Example}}{}
         \newcommand\citet\newcite
\newcommand\citep\cite

\newcommand\nl[1]{``\emph{#1}''}
\newcommand\wl[1]{\texttt{\footnotesize{#1}}}
\newcommand\den[1]{{\llbracket #1 \rrbracket}} \newcommand\vden[1]{{\llbracket #1 \rrbracket}_{\text{V}}}
\newcommand\score{\text{score}}

\newcommand\single{{\small\textsc{Single}}}
\newcommand\comp{{\small\textsc{Comp}}}

\newcommand\Traverse{\mathbb{T}}
\newcommand\Member{\mathbb{M}}

\title{Traversing Knowledge Graphs in Vector Space}

\author{Kelvin Guu \\
  Stanford University \\
  {\tt kguu@stanford.edu} \\\And
  John Miller \\
  Stanford University \\
  {\tt millerjp@stanford.edu} \\\And
  Percy Liang \\
  Stanford University \\
  {\tt pliang@cs.stanford.edu} \\}

\date{}

\begin{document}

\maketitle

\begin{abstract}

Path queries on a knowledge graph can be used to answer compositional questions such as
\nl{What languages are spoken by people living in Lisbon?}.
However, knowledge graphs often have missing facts (edges)
which disrupts path queries.
Recent models for knowledge base completion impute
missing facts by embedding knowledge graphs in vector spaces.
We show that these models can be recursively applied to answer path queries,
but that they suffer from cascading errors. This motivates a new ``compositional'' training objective,
which dramatically improves all models' ability to answer path queries, in some cases
more than doubling accuracy.
On a standard knowledge base completion task, we also demonstrate that compositional training
acts as a novel form of structural regularization,
reliably improving performance across all base models (reducing errors by up to 43\%) and achieving new state-of-the-art results.

\end{abstract}
 \section{Introduction}

Broad-coverage knowledge bases such as Freebase \citep{bollacker2008freebase}
support a rich array of reasoning and question answering applications,
but they are known to suffer from incomplete coverage \citep{min2013distant}.
For example, as of May 2015, Freebase has an entity Tad Lincoln (Abraham Lincoln's son), but does not have his ethnicity.
An elegant solution to incompleteness is
using vector space representations:~Controlling the dimensionality of the
vector space forces generalization to new facts
\citep{nickel2011three,nickel12yago,socher2013reasoning,riedel13universal,neelakantan2015compositional}.
In the example, we would hope to infer Tad's ethnicity from the ethnicity of his parents.

However, what is missing from these vector space models
is the original strength of knowledge bases: the ability to support
compositional queries \citep{ullman1985implementation}.
For example, we might ask what the ethnicity of Abraham Lincoln's daughter would be.
This can be formulated as a path query on the knowledge graph, and
we would like a method that can answer this efficiently,
while generalizing over missing facts and even missing or hypothetical entities
(Abraham Lincoln did not in fact have a daughter).

\begin{figure}
\begin{centering}
\includegraphics[scale=0.3]{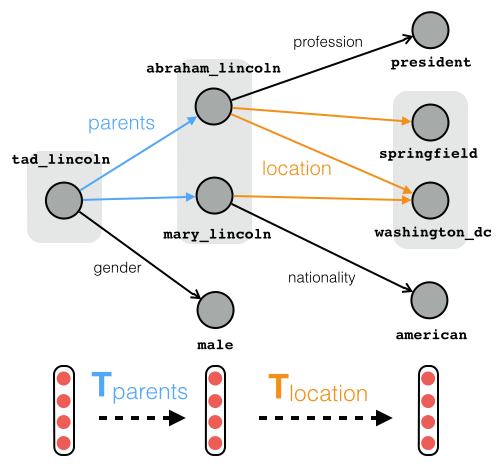}
\par\end{centering}
\protect\caption{
  \label{fig:main}
We propose performing path queries such as \wl{tad\_lincoln/parents/location}
(\nl{Where are Tad Lincoln's parents located?})
in a parallel low-dimensional vector space.
Here, entity sets (boxed) are represented as real vectors,
and edge traversal is driven by vector-to-vector transformations (e.g., matrix
multiplication).
}
\end{figure}

In this paper, we present a scheme to answer path queries on knowledge bases
by ``compositionalizing'' a broad class of vector space models that have been
used for knowledge base completion (see \reffig{main}).
At a high level, we interpret the base vector space model as implementing a soft edge traversal operator.
This operator can then be recursively applied to predict paths.
Our interpretation suggests a new \emph{compositional training} objective
that encourages better modeling of paths.
Our technique is applicable to a broad class of \emph{composable} models
that includes the bilinear model \citep{nickel2011three} and TransE \citep{bordes2013translating}.

We have two key empirical findings:
First, we show that compositional training enables us to answer path queries up
to at least length 5 by substantially reducing cascading errors present
in the base vector space model.
Second, we find that somewhat surprisingly, compositional training also
improves upon state-of-the-art performance for knowledge base completion,
which is a special case of answering unit length path queries.
Therefore, compositional training can also be seen as a new form of
structural regularization for existing models.

 \section{Task}
\label{sec:task}
We now give a formal definition of the task of answering path queries on a knowledge base.
Let $\mathcal{E}$ be a set of entities and $\mathcal{R}$ be a set
of binary relations. A knowledge graph $\mathcal{G}$ is defined as
a set of triples of the form $\left(s,r,t\right)$ where $s,t\in\mathcal{E}$
and $r\in\mathcal{R}$.
An example of a triple in Freebase is $(\wl{tad\_lincoln},\wl{parents},\wl{abraham\_lincoln})$.

A path query $q$ consists of an initial \emph{anchor entity,} $s$, followed
by a sequence of relations to be traversed,
$p = (r_{1},\ldots,r_{k})$.
The answer or denotation of the query, $\den{q}$, is the set of all entities that
can be reached from $s$ by traversing $p$.
Formally, this can be defined recursively:
\begin{align}
  \den{s} &\eqdef \{s\}, \label{eqn:denotation1}\\
  \den{q/r} &\eqdef \pc{ t : \exists s \in \den{q}, (s,r,t) \in \sG}. \label{eqn:denotation2}
\end{align}
For example, \wl{tad\_lincoln/parents/location} is a query $q$ that asks:
\nl{Where did Tad Lincoln's parents live?}.

For evaluation (see \refsec{results} for details), we
define the set of candidate answers to a query $\mathcal{C}(q)$ as the set
of all entities that ``type match'',
namely those that participate in the final relation of $q$ at least once;
and let $\sN(q)$ be the incorrect answers:
\begin{align}
\sC\left(s/r_1/ \cdots /r_k\right) &\eqdef \left\{ t\mid\exists e, \left(e,r_{k},t\right)\in\mathcal{G}\right\} \\
\mathcal{N}\left(q\right) &\eqdef \mathcal{C}\left(q\right)\backslash\llbracket q\rrbracket.
\end{align}

\paragraph{Knowledge base completion.} Knowledge base completion (KBC) is the task of predicting
whether a given edge $(s, r, t)$ belongs in the graph or not.
This can be formulated as a path query $q=s/r$ with candidate answer $t$.
 \section{Compositionalization}
\label{sec:compositionalize}

In this section, we show how to compositionalize existing KBC models to
answer path queries.
We start with a motivating example in \refsec{comp_example}, then
present the general technique in \refsec{comp_general}. This suggests a new
compositional training objective, described in \refsec{comp_training}.
Finally, we illustrate the technique for several more models in \refsec{comp_more}, which
we use in our experiments.

\subsection{Example} \label{sec:comp_example}
A common vector space model for knowledge base completion is the \emph{bilinear
model} \citep{nickel2011three}.
In this model,
we learn a vector $x_e \in \R^d$ for each entity $e \in \sE$
and a matrix $W_r \in \R^{d \times d}$
for each relation $r \in \sR$.
Given a query $s/r$ (asking for the set of entities connected to $s$ via relation $r$),
the bilinear model scores how likely $t \in \den{s/r}$ holds using
\begin{align} \label{eqn:bilinear}
	\mathrm{score}(s/r, t) = x_s^\top W_{r} x_t.
\end{align}

To motivate our compositionalization technique,
take $d = |\sE|$ and
suppose $W_r$ is the adjacency matrix for relation $r$
and entity vector $x_e$ is the indicator vector with a 1 in the entry corresponding to entity $e$.
Then, to answer a path query $q=s/r_1/\ldots/r_k$, we would then compute
\begin{align} \label{eqn:bilinear_comp}
\score(q, t) &= x_s^\top W_{r_{1}} \ldots W_{r_{k}} x_t.
\end{align}
It is easy to verify that the score counts the number of unique paths between $s$ and $t$
following relations $r_1/\ldots/r_k$.
Hence, any $t$ with positive score is a correct answer
($\den{q} = \{ t : \score(q,t) > 0 \}$).

Let us interpret \refeqn{bilinear_comp} recursively.
The model begins with an entity vector $x_s$, and sequentially applies
\emph{traversal operators} $\Traverse_{r_i}(v) = v^\top W_{r_i}$ for each $r_i$. Each
traversal operation results in a new ``set vector'' representing the entities
reached at that point in traversal (corresponding to the nonzero entries of the
set vector). Finally, it applies the \emph{membership operator} $\Member(v, x_t) =
v^\top x_t$ to check if $t \in \den{s/r_1/\ldots/r_k}$.
Writing graph traversal in this way immediately
suggests a useful generalization:
take $d$ much smaller than $|\mathcal{E}|$
and learn the parameters $W_r$ and $x_e$.

\subsection{General technique} \label{sec:comp_general}

The strategy used to extend the bilinear model of \refeqn{bilinear} to
the compositional model in \refeqn{bilinear_comp}
can be applied to any \emph{composable} model: namely, one that has a scoring function of the form:
\begin{align}
  \label{eqn:member_traverse}
	\mathrm{score}(s/r, t) = \Member(\Traverse_r(x_s), x_t)
\end{align}
for some choice of membership operator $\Member: \R^{d} \times \R^{d} \to \R$
and traversal operator $\Traverse_r:\R^{d} \to \R^{d}$.

We can now define the \emph{vector denotation} of a query $\vden{q}$
analogous to the definition of $\den{q}$ in \refeqn{denotation1} and \refeqn{denotation2}:
\begin{align}
  \vden{s} &\eqdef x_s, \\
  \vden{q/r}  &\eqdef \Traverse_{r}\left(\vden{q}\right).
\end{align}

The score function for a compositionalized model is then
\begin{align}
  \label{comp_score}
	\mathrm{score}(q, t) = \Member(\vden{q}, \vden{t}).
\end{align}
We would like $\vden{q}$ to approximately represent the set $\den{q}$ in
the sense that for every $e \in \den{q}$,
$\Member\left(\vden{q}, \vden{e}\right)$ is larger
than the values for $e \not\in \den{q}$.
Of course it is not possible to represent all sets perfectly, but in the next
section, we present a training objective that explicitly optimizes
$\Traverse$ and $\Member$ to
preserve path information.

\subsection{Compositional training} \label{sec:comp_training}

The score function in (\ref{comp_score}) naturally suggests a new compositional
training objective.
Let $\{(q_i, t_i)\}_{i=1}^N$ denote a set of path query training examples
with path lengths ranging from 1 to $L$.
We minimize the following max-margin objective:
\begin{align*}
	J(\Theta) = \sum_{i =1}^N \sum_{t' \in \mathcal{N}(q_i)} \left[1 - \text{margin}(q_i, t_i, t')\right]_{+}, \\
    \text{margin}(q, t, t') = \score(q, t) - \score(q, t'),
\end{align*}
where the parameters are the membership operator, the traversal operators, and
the entity vectors:
\begin{align*}
\small
  \Theta &= \left\{ \Member \right\} \cup \left\{ \Traverse_{r}:r\in\mathcal{R}\right\} \cup \left\{ x_{e} \in \R^d : e\in\mathcal{E}\right\}.
\end{align*}

This objective encourages the construction of ``set vectors'':
because there are path queries of different lengths and types, the model must
learn to produce an accurate set vector $\vden{q}$ after any sequence of
traversals. Another perspective is that each traversal operator is trained such that
its transformation preserves information in the set vector which might be needed in subsequent traversal steps.

In contrast, previously proposed training objectives for knowledge base completion
only train on queries of path length 1.
We will refer to this special case as \emph{single-edge training}.

In \refsec{results}, we show that compositional training
leads to substantially better results for both path query answering
and knowledge base completion. In \refsec{analysis}, we provide insight
into why.

\subsection{Other composable models} \label{sec:comp_more}
There are many possible candidates for $\Traverse$ and $\Member$.
For example, $\Traverse$ could be one's favorite neural network mapping from $\R^d$ to $\R^d$.
Here, we focus on two composable models that were both recently shown to
achieve state-of-the-art performance on knowledge base completion.

\paragraph{TransE.}
The TransE model of \citet{bordes2013translating} uses the scoring function
\begin{align}
	\score(s/r, t) = -\| x_{s}  + w_{r} - x_{t} \|_2^2.
\end{align}
where $x_s$, $w_r$ and $x_t$ are all $d$-dimensional vectors.

In this case, the model can be expressed using membership operator
\begin{align}
	\Member(v, x_t) = -\|v - x_t \|_2^2
\end{align}
and traversal operator $\Traverse_r(x_s) = x_s + w_r$. Hence, TransE can handle a path query $q=s/r_1/r_2/\cdots/r_k$ using
\begin{align*}
	\score(q, t) = -\|x_s + w_{r_1} + \cdots + w_{r_k} - x_t \|_2^2.
\end{align*}
We visualize the compositional TransE model in \reffig{cascading}.

\paragraph{Bilinear-Diag.}
The Bilinear-Diag model of \citet{yang2015embeddings} is a special case of the bilinear model with the relation matrices constrained to be diagonal. Alternatively, the model can be viewed as a variant of TransE with multiplicative interactions between entity and relation vectors.

\paragraph{Not all models can be compositionalized.}
It is important to point out that some models
are not naturally composable---for example, the
latent feature model of \citet{riedel13universal} and the neural
tensor network of \citet{socher2013reasoning}.
These approaches have scoring functions which combine $s$, $r$ and $t$ in a way
that does not involve an intermediate vector representing $s/r$ alone without $t$,
so they do not decompose according to \refeqn{member_traverse}.

\subsection{Implementation}
We use AdaGrad \citep{duchi10adagrad} to optimize $J(\Theta)$, which is in general non-convex.
Initialization scale, mini-batch size and step size were cross-validated
for all models. We initialize all parameters with i.i.d. Gaussians of variance
0.1 in every entry, use a mini-batch size of 300 examples, and a step size in
$[0.001, 0.1]$ (chosen via cross-validation) for all of the models. For each
example $q$, we sample 10 negative entities $t' \in \mathcal{N}(q)$. During
training, all of the entity vectors are constrained to lie on the unit ball,
and we clipped the gradients to the median of the observed gradients if the
update exceeded 3 times the median.

We first train on path queries of length 1 until convergence
and then train on all path queries until convergence.
This guarantees that the model masters basic edges
before composing them to form paths.
When training on path queries, we explicitly parameterize inverse relations.
For the bilinear model, we initialize $W_{r^{-1}}$ with $W_{r}^\top$. For TransE, we initialize $w_{r^{-1}}$ with $-w_r$. For Bilinear-Diag, we found initializing $w_{r^{-1}}$ with the exact inverse $1/w_{r}$ is numerically unstable, so we instead randomly initialize $w_{r^{-1}}$ with i.i.d Gaussians of variance 0.1 in every entry.
Additionally, for the bilinear model, we replaced the sum over $\mathcal{N}(q_i)$
in the objective with a max since it yielded slightly higher accuracy.
Our models are implemented using Theano \citep{bastien2012theano,bergstra2010scipy}.

 \section{Datasets}

In \refsec{baseDataset}, we describe two standard knowledge base completion datasets.
These consist of single-edge queries, so we call them \emph{base datasets}.
In \refsec{pathDataset}, we generate path query
datasets from these base datasets.

\begin{table}
\begin{centering}
\begin{tabular}{cc|c|c}
\multicolumn{2}{c|}{} & \textbf{WordNet} & \textbf{Freebase}\tabularnewline
\hline
\multicolumn{2}{c|}{\textbf{Relations}} & 11 & 13\tabularnewline
\multicolumn{2}{c|}{\textbf{Entities}} & 38,696 & 75,043\tabularnewline
\hline
\multirow{2}{*}{\textbf{Base}} & Train & 112,581 & 316,232\tabularnewline
 & Test & 10,544 & 23,733\tabularnewline
\hline
\multirow{2}{*}{\textbf{Paths}} & Train & 2,129,539 & 6,266,058\tabularnewline
 & Test & 46,577 & 109,557\tabularnewline
\end{tabular}
\par\end{centering}
\protect\caption{\label{tab:datasets}WordNet and Freebase statistics for base and path query datasets.}
\end{table}

\subsection{Base datasets}
\label{sec:baseDataset}
Our experiments are conducted using the subsets of WordNet and Freebase
from \citet{socher2013reasoning}. The statistics of these datasets and
their splits are given in \reftab{datasets}.

The WordNet and Freebase subsets exhibit substantial
differences that can influence model performance.
The Freebase subset is almost bipartite with most of the edges
taking the form $(s, r, t)$ for some \emph{person} $s$, relation $r$
and property $t$. In WordNet, both the source and target entities
are arbitrary words.

Both the raw WordNet and Freebase contain many relations that are almost
perfectly correlated with an inverse relation. For example, WordNet contains
both \wl{has\_part} and \wl{part\_of}, and Freebase contains both
\wl{parents} and \wl{children}. At test time, a query on an edge $(s, r,
t)$ is easy to answer if the inverse triple $(t, r^{-1}, s)$ was observed in the training
set. Following \citet{socher2013reasoning}, we account for this by excluding such
``trivial'' queries from the test set.

\subsection{Path query datasets}
\label{sec:pathDataset}

Given a base knowledge graph, we generate path queries by performing random walks on the graph.
If we view compositional training as a form of regularization, this approach
allows us to generate extremely large amounts of auxiliary training data.
The procedure is given below.

Let $\mathcal{G}_{\mathrm{train}}$ be the training graph, which consists only of
the edges in the training set of the base dataset.
We then repeatedly generate training examples with the following procedure:
\begin{enumerate}
\item Uniformly sample a path length $L \in \{1, \dots, L_{\mathrm{max}}\}$, and uniformly sample
a starting entity $s \in \mathcal{E}$.
\item Perform a random walk beginning at entity $s$ and continuing $L$ steps.
\begin{enumerate}
\item At step $i$ of the walk, choose a relation $r_i$ uniformly from the set
of relations incident on the current entity $e$.
\item Choose the next entity uniformly from the set of entities reachable via $r_i$.
\end{enumerate}
\item Output a query-answer pair, $(q,t)$, where $q=s/r_1/\cdots/r_L$ and $t$ is the final entity of the random walk.
\end{enumerate}
In practice, we do not sample paths of length 1 and instead directly add all of
the edges from $\mathcal{G}_{\mathrm{train}}$ to the path query dataset.

To generate a path query test set, we repeat the above procedure except using
the graph $\mathcal{G}_{\mathrm{full}}$, which is
$\mathcal{G}_{\mathrm{train}}$ plus all of the test edges from the base
dataset. Then we remove any queries from the test set
that also appeared in the training set. The statistics for the path query
datasets are presented in \reftab{datasets}.

\section{Main results} \label{sec:results}

We evaluate the models derived in \refsec{compositionalize} on two tasks: path query answering
and knowledge base completion.
On both tasks, we show that the compositional training strategy proposed in \refsec{comp_training}
leads to substantial performance gains over standard single-edge training.
We also compare directly against the KBC results of \citet{socher2013reasoning},
demonstrating that previously inferior models now match or outperform state-of-the-art
models after compositional training.

\paragraph{Evaluation metric.}
Numerous metrics have been used to evaluate knowledge base queries, including
hits at 10 (percentage of correct answers ranked in the top 10) and mean rank.
We evaluate on hits at 10, as well as a normalized version
of mean rank, \emph{mean quantile}, which better accounts for the total number of candidates.
For a query $q$, the quantile of a correct answer $t$ is
the fraction of incorrect answers ranked after $t$:
\begin{align}
  \small
\frac{\left|\left\{ t^{\prime}\in\mathcal{N}\left(q\right) : \score(q,t^{\prime})<\score(q,t)\right\} \right|}{\left|\mathcal{N}\left(q\right)\right|}
\end{align}
The quantile ranges from 0 to 1, with 1 being optimal.
Mean quantile is then defined to be the average quantile score over all examples in the dataset.
To illustrate why normalization is important, consider a set of queries on the
relation \wl{gender}. A model that predicts the incorrect gender on every query
would receive a mean rank of 2 (since there are only 2 candidate answers),
which is fairly good in absolute terms,
whereas the mean quantile would be 0,
rightfully penalizing the model.

As a final note, several of the queries in the Freebase path dataset are
``type-match trivial'' in the sense that all of the type matching candidates
$\mathcal{C}(q)$ are correct answers to the query. In this case, mean quantile
is undefined and we exclude such queries from evaluation.

\begin{table*}[t]
\begin{centering}
\begin{tabular}{cc|ccc|ccc|ccc}
\multicolumn{2}{c|}{} & \multicolumn{3}{c|}{\textbf{Bilinear}} & \multicolumn{3}{c|}{\textbf{Bilinear-Diag}} & \multicolumn{3}{c}{\textbf{TransE}}\tabularnewline
\hline
\multicolumn{2}{c|}{\textbf{Path query task}} & $\single$ & $\comp$ & (\%red) & $\single$ & $\comp$ & (\%red) & $\single$ & $\comp$ & (\%red)\tabularnewline
\hline
\multirow{2}{*}{WordNet} & MQ & 84.7 & 89.4 & \textbf{30.7} & 59.7 & 90.4 & \textbf{76.2} & 83.7 & 93.3 & \textbf{58.9}\tabularnewline
 & H@10 & 43.6 & 54.3 & \textbf{19.0} & 7.9 & 31.1 & \textbf{25.4} & 13.8 & 43.5 & \textbf{34.5}\tabularnewline
\multirow{2}{*}{Freebase} & MQ & 58.0 & 83.5 & \textbf{60.7} & 57.9 & 84.8 & \textbf{63.9} & 86.2 & 88 & \textbf{13.0}\tabularnewline
 & H@10 & 25.9 & 42.1 & \textbf{21.9} & 23.1 & 38.6 & \textbf{20.2} & 45.4 & 50.5 & \textbf{9.3}\tabularnewline
\hline
\multicolumn{2}{c|}{\textbf{KBC task}} & $\single$ & $\comp$ & (\%red) & $\single$ & $\comp$ & (\%red) & $\single$ & $\comp$ & (\%red)\tabularnewline
\hline
\multirow{2}{*}{WordNet} & MQ & 76.1 & 82.0 & \textbf{24.7} & 76.5 & 84.3 & \textbf{33.2} & 75.5 & 86.1 & \textbf{43.3}\tabularnewline
 & H@10 & 19.2 & 27.3 & \textbf{10.0} & 12.9 & 14.4 & \textbf{1.72} & 4.6 & 16.5 & \textbf{12.5}\tabularnewline
\multirow{2}{*}{Freebase} & MQ & 85.3 & 91.0 & \textbf{38.8} & 84.6 & 89.1 & \textbf{29.2} & 92.7 & 92.8 & \textbf{1.37}\tabularnewline
 & H@10 & 70.2 & 76.4 & \textbf{20.8} & 63.2 & 67.0 & \textbf{10.3} & 78.8 & 78.6 & -0.9\tabularnewline
\end{tabular}
\par\end{centering}
\protect\caption{\label{tab:results}\textbf{Path query answering and knowledge base completion.} We compare
the performance of single-edge training ($\single$) vs compositional
training ($\comp$). MQ: mean quantile, H@10: hits at 10, \%red: percentage
reduction in error.}
\end{table*}

\paragraph{Overview.}
The upper half of \reftab{results} shows that compositional training improves path querying performance across
all models and metrics on both datasets, reducing error by up to $76.2\%$.

The lower half of \reftab{results} shows that surprisingly, compositional training also improves performance on
knowledge base completion across almost all models, metrics and datasets.
On WordNet, TransE benefits the most, with a $43.3\%$ reduction in error.
On Freebase, Bilinear benefits the most, with a $38.8\%$ reduction in error.

In terms of mean quantile, the best overall model is TransE ($\comp$).
In terms of hits at 10, the best model on WordNet is Bilinear ($\comp$), while the best
model on Freebase is TransE ($\comp$).

\paragraph{Deduction and Induction.}
\reftab{ded_ind} takes a deeper look at performance on path query answering.
We divided path queries into two subsets: \emph{deduction} and \emph{induction}.
The \emph{deduction} subset consists of queries $q=s/p$ where the source and target entities $\den{q}$ are connected via relations $p$ in the training graph $\mathcal{G}_{\mathrm{train}}$, but the specific query $q$ was never seen during training. Such queries can be answered by performing explicit traversal on the training graph, so this subset tests a model's ability to approximate the underlying training graph and predict the existence of a path from a collection of single edges.
The \emph{induction} subset consists of all other queries. This means that
at least one edge was missing on all paths following $p$ from source to target in the
training graph. Hence, this subset tests a model's generalization ability
and its robustness to missing edges.

Performance on the \emph{deduction} subset of the dataset is disappointingly low
for models trained with single-edge training: they struggle to answer path queries
even when \emph{all edges in the path query have been seen at training time.}
Compositional training dramatically reduces these errors, sometimes doubling mean quantile.
In \refsec{analysis}, we analyze how this might be possible. After compositional training, performance on the harder \emph{induction} subset
is also much stronger. Even when edges are missing along a path, the models are able to
infer them.

\begin{table}[t]
\begin{centering}
\begin{tabular}{cc|cc|cc}
\multicolumn{2}{c|}{\textbf{Path query task}} & \multicolumn{2}{c|}{\textbf{WordNet}} & \multicolumn{2}{c}{\textbf{Freebase}}\tabularnewline
\multicolumn{2}{c|}{} & Ded. & Ind. & Ded. & Ind.\tabularnewline
\hline
\multirow{2}{*}{\textbf{Bilinear}} & $\single$ & 96.9 & 66.0 & 49.3 & 49.4\tabularnewline
 & $\comp$ & \textbf{98.9} & \textbf{75.6} & \textbf{82.1} & \textbf{70.6}\tabularnewline
\hline
\multirow{2}{*}{\textbf{Bi-Diag}} & $\single$ & 56.3 & 51.6 & 49.3 & 50.2\tabularnewline
 & $\comp$ & \textbf{98.5} & \textbf{78.2} & \textbf{84.5} & \textbf{72.8}\tabularnewline
\hline
\multirow{2}{*}{\textbf{TransE}} & $\single$ & 92.6 & 71.7 & 85.3 & 72.4\tabularnewline
 & $\comp$ & \textbf{99.0} & \textbf{87.4} & \textbf{87.5} & \textbf{76.3}\tabularnewline
\end{tabular}
\par\end{centering}
\protect\caption{\label{tab:ded_ind}\textbf{Deduction and induction.}
We compare mean quantile performance of single-edge training ($\single$) vs compositional training ($\comp$).
Length 1 queries are excluded.}
\end{table}

\paragraph{Interpretable queries.}
Although our path datasets consists of random queries, both datasets contain a large number of useful, interpretable queries. Results on a few illustrative examples are shown in \reftab{interpretable}.

\begin{table*}[t]
\begin{centering}
\begin{tabular}{cc|cc}
\multicolumn{2}{c|}{\textbf{Interpretable Queries}} & Bilinear $\single$ &  Bilinear $\comp$  \tabularnewline
\hline
\multirow{1}{*}{\wl{X/institution/institution$^{-1}$/profession}} & & 50.0 & \textbf{93.6} \tabularnewline
\multirow{1}{*}{\wl{X/parents/religion}} & & 81.9 & \textbf{97.1}  \tabularnewline
\multirow{1}{*}{\wl{X/nationality/nationality$^{-1}$/ethnicity$^{-1}$}} & & 68.0 & \textbf{87.0} \tabularnewline
\multirow{1}{*}{\wl{X/has\_part/has\_instance$^{-1}$}} & & 92.6 & \textbf{95.1} \tabularnewline
\multirow{1}{*}{\wl{X/type\_of/type\_of/type\_of}} & & 72.8 & \textbf{79.4} \tabularnewline
\end{tabular}
\par\end{centering}

\protect\caption{\label{tab:interpretable}Path query performance (mean quantile) on a selection of interpretable queries. We compare Bilinear $\single$ and Bilinear $\comp$. Meanings of each query (descending): ``What professions are there at X's institution?''; ``What is the religion of X's parents?''; ``What are the ethnicities of people from the same country as X?''; ``What types of parts does X have?''; and the transitive ``What is X a type of?''.  (Note that a relation $r$ and its inverse $r^{-1}$ do not necessarily cancel out if $r$ is not a one-to-one mapping. For example, \wl{X/institution/institution$^{-1}$} denotes the set of all people who work at the institution X works at, which is not just X.)}
\end{table*}

\paragraph{Comparison with \citet{socher2013reasoning}.}

Here, we measure performance on the KBC task in terms of the accuracy metric of \citet{socher2013reasoning}.
This evaluation involves sampled negatives, and is hence noisier than mean quantile,
but makes our results directly comparable to \citet{socher2013reasoning}.
Our results show that previously inferior models such as the bilinear model can outperform state-of-the-art
models after compositional training.

\citet{socher2013reasoning} proposed parametrizing each entity vector as the average
of vectors of words in the entity ($w_{\wl{tad\_lincoln}} = \frac12 (w_{\wl{tad}} + w_{\wl{lincoln}})$,
and pretraining these word vectors using the method of \citet{turian2010word}.
\reftab{socher_compare} reports results
when using this approach in conjunction with compositional training.
We initialized all models with word vectors from \citet{pennington2014glove}.
We found that compositionally trained models outperform the neural tensor network (NTN) on WordNet,
while being only slightly behind on Freebase.
(We did not use word vectors in any of our other experiments.)

When the strategy of averaging word vectors to form entity vectors is not applied,
our compositional models are significantly better on WordNet and slightly better on Freebase.
It is worth noting that in many domains, entity names are not lexically meaningful, so
word vector averaging is not always meaningful.

\begin{table}
\begin{centering}
\begin{tabular}{c|cc|cc}
Accuracy & \multicolumn{2}{c|}{\textbf{WordNet}} & \multicolumn{2}{c}{\textbf{Freebase}}\tabularnewline
 & EV & WV & EV & WV\tabularnewline
\hline
NTN & 70.6 & 86.2 & 87.2 & \textbf{90.0}\tabularnewline
\hline
Bilinear \comp & 77.6 & \textbf{87.6} & 86.1 & 89.4\tabularnewline
TransE \comp & \textbf{80.3} & 84.9 & \textbf{87.6} & 89.6\tabularnewline
\end{tabular}
\par\end{centering}

\protect\caption{\label{tab:socher_compare}Model performance in terms of accuracy.
  EV: entity vectors are separate (initialized randomly);
  WV: entity vectors are average of word vectors (initialized with pretrained word vectors).}
\end{table}

 \section{Analysis} \label{sec:analysis}

In this section, we try to understand why
compositional training is effective.
For concreteness, everything is
described in terms of the bilinear model. We will refer to the compositionally trained model as $\comp$, and the
model trained with single-edge training as $\single$.

\subsection{Why does compositional training improve path query answering?}

\begin{figure}
\begin{centering}
\includegraphics[scale=0.5]{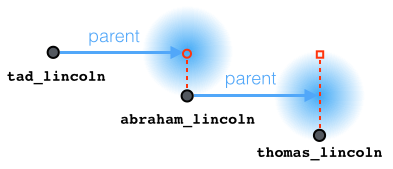}
\par\end{centering}

\protect\caption{\label{fig:cascading} \textbf{Cascading errors visualized for TransE.}
Each node represents the position of an entity in vector space. The relation $\wl{parent}$ is
ideally a simple horizontal translation, but each traversal introduces noise. The red circle
is where we expect Tad's parent to be. The red square is where we expect Tad's grandparent to be.
Dotted red lines show that error grows larger as we traverse farther away from Tad.
Compositional training pulls the entity vectors closer to the ideal arrangement.
}
\end{figure}

It is tempting to think that if $\single$
has accurately modeled individual edges in a graph, it should accurately
model the paths that result from those edges. This intuition turns
out to be incorrect, as revealed by $\single$'s relatively
weak performance on the path query dataset.
We hypothesize that this is due to cascading errors
along the path.
For a given edge $\left(s,r,t\right)$ on the path, single-edge training
encourages $x_{t}$ to be closer to $x_{s}^{\top} W_{r}$
than any other incorrect $x_{t^{\prime}}$. However, once this is
achieved by a margin of $1$, it does not push $x_{t}$ any closer
to $x_{s}^{\top}W_{r}$. The remaining discrepancy is noise which gets
added at each step of path traversal. This is illustrated schematically in \reffig{cascading}.

\begin{figure}
\begin{centering}
\includegraphics[scale=0.25]{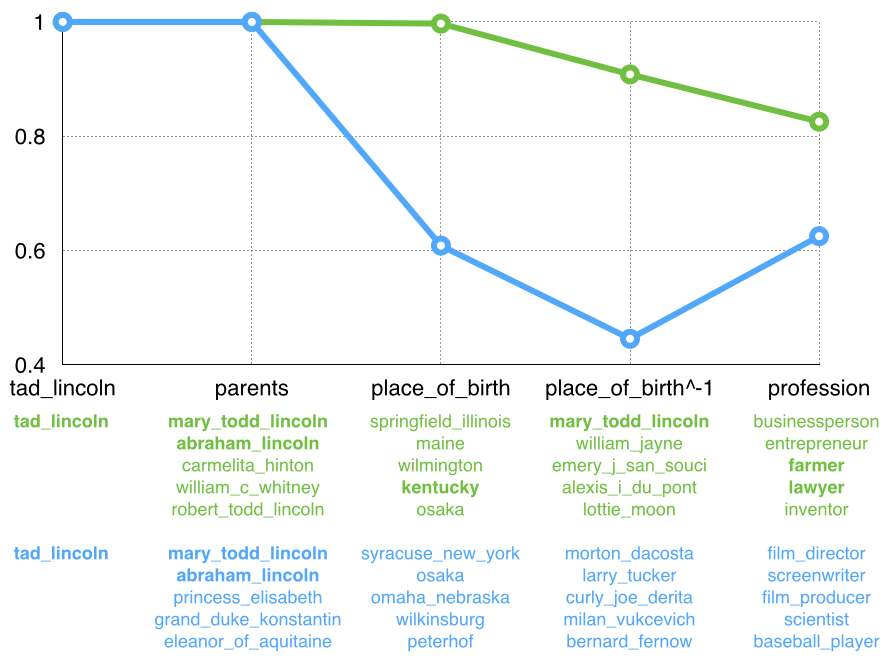}
\par\end{centering}

\protect\caption{\label{fig:RQ} Reconstruction quality (RQ) at
each step of the query
\wl{tad\_lincoln/parents/place\_of\_birth/}
\wl{place\_of\_birth$^{-1}$/profession}.
$\comp$ experiences significantly less degradation in RQ as
path length increases.  Correspondingly, the set of 5 highest scoring entities
computed at each step using $\comp$ (green) is
significiantly more accurate than the set given by $\single$ (blue). Correct
entities are bolded.}
\end{figure}

To observe this phenomenon empirically, we examine how well
a model handles each \emph{intermediate} step of a path query. We can do this by measuring
the \emph{reconstruction quality} (RQ) of the set vector produced after each traversal operation.
Since each intermediate stage is itself a valid path query, we define RQ to be
the average quantile over all entities that belong in $\den{q}$:
\begin{align}
  \text{RQ}\left(q\right)=\frac{1}{\left|\llbracket q\rrbracket\right|}\sum_{t\in\llbracket q\rrbracket}\mbox{quantile}\left(q, t\right)
\end{align}
When all
entities in $\den{q}$ are ranked above all incorrect entities, RQ is 1.
In \reffig{RQ}, we illustrate how RQ changes over the course of a query.

Given the nature of cascading errors, it might seem reasonable to address the
problem by adding a term to our objective which explicitly encourages $x_s^\top
W_r$ to be as close as possible to $x_t$.  With this motivation, we tried adding
$\lambda \|x_s^\top W_r - x_t\|_2^2$ term to the objective of the bilinear
model and a $\lambda \|x_s + w_r - x_t\|_2^2$ term to the objective of TransE.
We experimented with different settings of $\lambda$ over the range $[0.001, 100]$.
In no case did this additional $\ell_2$ term improve $\single$'s performance on the
path query or single edge dataset. These results suggest that compositional
training is a more effective way to combat cascading errors.

\subsection{Why does compositional training improve knowledge base completion?}

\reftab{results} reveals that $\comp$ also performs
better on the single-edge task of knowledge base completion.
This is somewhat surprising, since $\single$ is trained on
a training set which distributionally matches the test set, whereas
$\comp$ is not.
However, $\comp$'s better performance on path queries suggests that
there must be another factor at play.
At a high level, training on paths must be providing some form of structural regularization which
reduces cascading errors.
Indeed, paths in a knowledge graph have proven to be important features for
predicting the existence of single edges \citep{lao2011pathranking,neelakantan2015compositional}. For example,
consider the following Horn clause:
\[
  \small
\mbox{parents}\left(x,y\right)\wedge\mbox{location}\left(y,z\right)
\Rightarrow\mbox{place\_of\_birth}\left(x,z\right),
\]
which states that if $x$ has a parent with location $z$, then $x$
has place of birth $z$. The body of the Horn clause expresses a
path from $x$ to $z$.
If $\comp$ models the path better, then it should be better able to use that
knowledge to infer the head of the Horn clause.

More generally, consider Horn clauses of the form $p\Rightarrow r$,
where $p=r_{1}/\ldots/r_{k}$ is a path type and $r$ is the relation being
predicted. Let us focus on Horn clauses with high precision as defined by:
\begin{align}
  \text{prec}(p)=\frac{\left|\llbracket p\rrbracket\cap\llbracket r\rrbracket\right|}{\left|\llbracket p\rrbracket\right|},
\end{align}
where $\den{p}$ is the set of entity pairs connected by $p$, and similarly for $\den{r}$.

Intuitively, one way for the model to implicitly learn and exploit such a Horn
clause would be to satisfy the following two criteria:
\begin{enumerate}
\item The model should ensure a \emph{consistent} spatial relationship
between entity pairs that are related by the path type $p$; that is,
keeping $x_{s}^{\top}W_{r_{1}}\ldots W_{r_{k}}$
close to $x_{t}$ for \emph{all} valid $\left(s,t\right)$ pairs.
\item The model's representation of the path type $p$ and relation $r$
  should capture that spatial relationship; that is,
  $x_{s}^{\top}W_{r_{1}}\ldots W_{r_{k}}\approx x_{t}$ implies $x_{s}^{\top}W_{r}\approx x_{t}$,
  or simply $W_{r_{1}}\ldots W_{r_{k}}\approx W_{r}$.
\end{enumerate}
We have already seen empirically that $\single$ does not
meet criterion 1, because cascading errors cause it to put incorrect
entity vectors $x_{t^{\prime}}$ closer to $x_{s}^{\top}W_{r_{1}}\ldots W_{r_{k}}$
than the correct entity.  $\comp$ mitigates these errors.

To empirically verify that $\comp$ also
does a better job of meeting criterion 2,
we perform the following: for a
path type $p$ and relation $r$, define $\mbox{dist}(p, r)$ to be the angle between
their corresponding matrices (treated as vectors in $\R^{d^2}$).
This is a natural measure because $x_s^\top W_r x_t$ computes
the matrix inner product between $W_r$ and $x_s x_t^\top$. Hence, any
matrix with small distance from $W_r$ will produce nearly the same scores as $W_r$
for the same entity pairs.

\begin{figure}
\begin{centering}
\includegraphics[width=\linewidth]{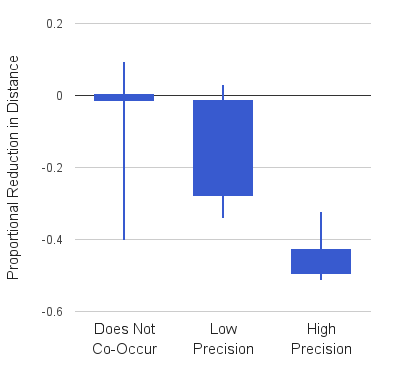}
\par\end{centering}
\protect\caption{\label{fig:angles} We divide paths of length 2 into high precision ($>$ 0.3), low precision ($\leq$ 0.3), and not co-occuring with $r$. Here $r=\mathrm{\wl{nationality}}$. Each box plot shows the min, max, and first and third quartiles of $\Delta \mbox{dist}(p, r)$. As hypothesized, compositional training results in large decreases in $\mbox{dist}(p, r)$ for high precision paths $p$, modest decreases for low precision paths, and little to no decreases for irrelevant paths.}

\end{figure}

If $\comp$ is better at capturing the correlation between $p$ and $r$, then we would expect that
when $\text{prec}(p)$ is high, compositional training should shrink $\text{dist}(p, r)$ more.
To confirm this hypothesis, we enumerated over all 676 possible paths of length 2 (including inverted relations), and
examined the proportional reduction in $\text{dist}(p, r)$ caused by compositional training,
\begin{align}
  \Delta\mbox{dist}(p, r)=\frac{\mbox{dist}_{\comp}(p, r)-\mbox{dist}_{\single}(p, r)}{\mbox{dist}_{\single}(p, r)}.
\end{align}
\reffig{angles} shows that higher precision paths indeed correspond to larger reductions
in $\text{dist}(p, r)$.

 \section{Related work}

\paragraph{Knowledge base completion with vector space models.}
Many models have been proposed for knowledge base completion, including those
reviewed in \refsec{comp_more} \citep{nickel2011three,bordes2013translating,yang2015embeddings,socher2013reasoning}.
\citet{dong2014knowledge} demonstrated that KBC models can improve the quality of relation extraction
by serving as graph-based priors.
\citet{riedel13universal} showed that such models can be also be directly used for open-domain relation extraction.
Our compositional training technique is an orthogonal improvement that could help any composable model.

\paragraph{Distributional compositional semantics.}

Previous works have explored compositional vector space representations in the
context of logic and sentence interpretation.
In \citet{socher2012mvrnn}, a matrix is associated with each word of a sentence, and can be used to recursively modify the meaning of nearby constituents.
\citet{grefenstette2013calculus} explored the ability of tensors to simulate logical calculi.
\citet{bowman2014recursive} showed that recursive neural networks can learn to distinguish important semantic relations.
\citet{socher2014grounded} found that compositional models were powerful enough to describe and retrieve images.

We demonstrate that compositional representations are also useful in the context of knowledge base
querying and completion. In the aforementioned work, compositional models produce vectors which represent truth values, sentiment or image features.
In our approach, vectors represent sets of entities constituting the denotation of a knowledge base query.

\paragraph{Path modeling.}

Numerous methods have been proposed to leverage path information
for knowledge base completion and question answering.
\citet{nickel2014reducing} proposed combining low-rank models with sparse path features.
\citet{lao2010relational} used random walks as features and \citet{gardner2014incorporating} extended this approach by using vector space similarity to govern random walk probabilities.
\citet{neelakantan2015compositional} addressed the problem of path sparsity by embedding paths using a recurrent neural network.
\citet{perozzi2014deepwalk} sampled random walks on social networks as training examples, with a different goal to classify nodes in the network.
\citet{bordes2014qa} embed paths as a sum of relation vectors for question answering.
Our approach is unique in modeling the denotation of each intermediate step of a path query, and using this information to regularize the spatial arrangement of entity vectors.

\section{Discussion}

We introduced the task of answering path queries on an incomplete knowledge base,
and presented a general technique for compositionalizing a broad class of vector
space models.
Our experiments show that compositional training leads to state-of-the-art
performance on both path query answering and knowledge base completion.

There are several key ideas from this paper:
regularization by augmenting the dataset with paths,
representing sets as low-dimensional vectors in a context-sensitive way,
and performing function composition using vectors.
We believe these three could all have greater applicability in the development of vector space models
for knowledge representation and inference.

\paragraph{Reproducibility}
Our code, data, and experiments are available on the CodaLab platform at \url{https://www.codalab.org/worksheets/0xfcace41fdeec45f3bc6ddf31107b829f}.

\paragraph{Acknowledgments}
We would like to thank Gabor Angeli for fruitful discussions and the anonymous reviewers for their valuable feedback.
We gratefully acknowledge the support of the Google Natural Language Understanding Focused Program
and the National Science Foundation Graduate Research Fellowship under Grant No. DGE-114747.

\bibliographystyle{acl}
\bibliography{all}

\begin{thebibliography}{}

\bibitem[\protect\citename{Bastien \bgroup et al.\egroup
  }2012]{bastien2012theano}
F.~Bastien, P.~Lamblin, R.~Pascanu, J.~Bergstra, I.~J. Goodfellow, A.~Bergeron,
  N.~Bouchard, and Y.~Bengio.
\newblock 2012.
\newblock Theano: new features and speed improvements.
\newblock Deep Learning and Unsupervised Feature Learning NIPS 2012 Workshop.

\bibitem[\protect\citename{Bergstra \bgroup et al.\egroup
  }2010]{bergstra2010scipy}
J.~Bergstra, O.~Breuleux, F.~Bastien, P.~Lamblin, R.~Pascanu, G.~Desjardins,
  J.~Turian, D.~Warde-Farley, and Y.~Bengio.
\newblock 2010.
\newblock Theano: a {CPU} and {GPU} math expression compiler.
\newblock In {\em Proceedings of the Python for Scientific Computing Conference
  ({SciPy})}.

\bibitem[\protect\citename{Bollacker \bgroup et al.\egroup
  }2008]{bollacker2008freebase}
K.~Bollacker, C.~Evans, P.~Paritosh, T.~Sturge, and J.~Taylor.
\newblock 2008.
\newblock {F}reebase: a collaboratively created graph database for structuring
  human knowledge.
\newblock In {\em International Conference on Management of Data (SIGMOD)},
  pages 1247--1250.

\bibitem[\protect\citename{Bordes \bgroup et al.\egroup
  }2013]{bordes2013translating}
A.~Bordes, N.~Usunier, A.~Garcia-Duran, J.~Weston, and O.~Yakhnenko.
\newblock 2013.
\newblock Translating embeddings for modeling multi-relational data.
\newblock In {\em Advances in Neural Information Processing Systems (NIPS)},
  pages 2787--2795.

\bibitem[\protect\citename{Bordes \bgroup et al.\egroup }2014]{bordes2014qa}
A.~Bordes, S.~Chopra, and J.~Weston.
\newblock 2014.
\newblock Question answering with subgraph embeddings.
\newblock In {\em Empirical Methods in Natural Language Processing (EMNLP)}.

\bibitem[\protect\citename{Bowman \bgroup et al.\egroup
  }2014]{bowman2014recursive}
S.~R. Bowman, C.~Potts, and C.~D. Manning.
\newblock 2014.
\newblock Can recursive neural tensor networks learn logical reasoning?
\newblock In {\em International Conference on Learning Representations (ICLR)}.

\bibitem[\protect\citename{Dong \bgroup et al.\egroup }2014]{dong2014knowledge}
X.~Dong, E.~Gabrilovich, G.~Heitz, W.~Horn, N.~Lao, K.~Murphy, T.~Strohmann,
  S.~Sun, and W.~Zhang.
\newblock 2014.
\newblock Knowledge {v}ault: A web-scale approach to probabilistic knowledge
  fusion.
\newblock In {\em International Conference on Knowledge Discovery and Data
  Mining (KDD)}, pages 601--610.

\bibitem[\protect\citename{Duchi \bgroup et al.\egroup }2010]{duchi10adagrad}
J.~Duchi, E.~Hazan, and Y.~Singer.
\newblock 2010.
\newblock Adaptive subgradient methods for online learning and stochastic
  optimization.
\newblock In {\em Conference on Learning Theory (COLT)}.

\bibitem[\protect\citename{Gardner \bgroup et al.\egroup
  }2014]{gardner2014incorporating}
M.~Gardner, P.~Talukdar, J.~Krishnamurthy, and T.~Mitchell.
\newblock 2014.
\newblock Incorporating vector space similarity in random walk inference over
  knowledge bases.
\newblock In {\em Empirical Methods in Natural Language Processing (EMNLP)}.

\bibitem[\protect\citename{Grefenstette}2013]{grefenstette2013calculus}
E.~Grefenstette.
\newblock 2013.
\newblock Towards a formal distributional semantics: Simulating logical calculi
  with tensors.
\newblock {\em arXiv preprint arXiv:1304.5823}.

\bibitem[\protect\citename{Lao and Cohen}2010]{lao2010relational}
N.~Lao and W.~W. Cohen.
\newblock 2010.
\newblock Relational retrieval using a combination of path-constrained random
  walks.
\newblock {\em Machine learning}, 81(1):53--67.

\bibitem[\protect\citename{Lao \bgroup et al.\egroup }2011]{lao2011pathranking}
N.~Lao, T.~Mitchell, and W.~W. Cohen.
\newblock 2011.
\newblock Random walk inference and learning in a large scale knowledge base.
\newblock In {\em Empirical Methods in Natural Language Processing (EMNLP)},
  pages 529--539.

\bibitem[\protect\citename{Min \bgroup et al.\egroup }2013]{min2013distant}
B.~Min, R.~Grishman, L.~Wan, C.~Wang, and D.~Gondek.
\newblock 2013.
\newblock Distant supervision for relation extraction with an incomplete
  knowledge base.
\newblock In {\em North American Association for Computational Linguistics
  (NAACL)}, pages 777--782.

\bibitem[\protect\citename{Neelakantan \bgroup et al.\egroup
  }2015]{neelakantan2015compositional}
A.~Neelakantan, B.~Roth, and A.~McCallum.
\newblock 2015.
\newblock Compositional vector space models for knowledge base completion.
\newblock In {\em Association for Computational Linguistics (ACL)}.

\bibitem[\protect\citename{Nickel \bgroup et al.\egroup }2011]{nickel2011three}
M.~Nickel, V.~Tresp, and H.~Kriegel.
\newblock 2011.
\newblock A three-way model for collective learning on multi-relational data.
\newblock In {\em International Conference on Machine Learning (ICML)}, pages
  809--816.

\bibitem[\protect\citename{Nickel \bgroup et al.\egroup }2012]{nickel12yago}
M.~Nickel, V.~Tresp, and H.~Kriegel.
\newblock 2012.
\newblock Factorizing {YAGO}.
\newblock In {\em World Wide Web (WWW)}.

\bibitem[\protect\citename{Nickel \bgroup et al.\egroup
  }2014]{nickel2014reducing}
M.~Nickel, X.~Jiang, and V.~Tresp.
\newblock 2014.
\newblock Reducing the rank in relational factorization models by including
  observable patterns.
\newblock In {\em Advances in Neural Information Processing Systems (NIPS)},
  pages 1179--1187.

\bibitem[\protect\citename{Pennington \bgroup et al.\egroup
  }2014]{pennington2014glove}
J.~Pennington, R.~Socher, and C.~D. Manning.
\newblock 2014.
\newblock Glove: Global vectors for word representation.
\newblock In {\em Empirical Methods in Natural Language Processing (EMNLP)}.

\bibitem[\protect\citename{Perozzi \bgroup et al.\egroup
  }2014]{perozzi2014deepwalk}
B.~Perozzi, R.~Al-Rfou, and S.~Skiena.
\newblock 2014.
\newblock Deepwalk: Online learning of social representations.
\newblock In {\em International Conference on Knowledge Discovery and Data
  Mining (KDD)}, pages 701--710.

\bibitem[\protect\citename{Riedel \bgroup et al.\egroup
  }2013]{riedel13universal}
S.~Riedel, L.~Yao, and A.~McCallum.
\newblock 2013.
\newblock Relation extraction with matrix factorization and universal schemas.
\newblock In {\em North American Association for Computational Linguistics
  (NAACL)}.

\bibitem[\protect\citename{Socher \bgroup et al.\egroup }2012]{socher2012mvrnn}
R.~Socher, B.~Huval, C.~D. Manning, and A.~Y. Ng.
\newblock 2012.
\newblock Semantic compositionality through recursive matrix-vector spaces.
\newblock In {\em Empirical Methods in Natural Language Processing and
  Computational Natural Language Learning (EMNLP/CoNLL)}, pages 1201--1211.

\bibitem[\protect\citename{Socher \bgroup et al.\egroup
  }2013]{socher2013reasoning}
R.~Socher, D.~Chen, C.~D. Manning, and A.~Ng.
\newblock 2013.
\newblock Reasoning with neural tensor networks for knowledge base completion.
\newblock In {\em Advances in Neural Information Processing Systems (NIPS)},
  pages 926--934.

\bibitem[\protect\citename{Socher \bgroup et al.\egroup
  }2014]{socher2014grounded}
R.~Socher, A.~Karpathy, Q.~V. Le, C.~D. Manning, and A.~Y. Ng.
\newblock 2014.
\newblock Grounded compositional semantics for finding and describing images
  with sentences.
\newblock {\em Transactions of the Association for Computational Linguistics
  (TACL)}, 2:207--218.

\bibitem[\protect\citename{Turian \bgroup et al.\egroup }2010]{turian2010word}
J.~Turian, L.~Ratinov, and Y.~Bengio.
\newblock 2010.
\newblock Word representations: a simple and general method for semi-supervised
  learning.
\newblock In {\em Proceedings of the 48th annual meeting of the association for
  computational linguistics}, pages 384--394.

\bibitem[\protect\citename{Ullman}1985]{ullman1985implementation}
J.~D. Ullman.
\newblock 1985.
\newblock Implementation of logical query languages for databases.
\newblock {\em ACM Transactions on Database Systems (TODS)}, 10(3):289--321.

\bibitem[\protect\citename{Yang \bgroup et al.\egroup
  }2015]{yang2015embeddings}
B.~Yang, W.~Yih, X.~He, J.~Gao, and L.~Deng.
\newblock 2015.
\newblock Embedding entities and relations for learning and inference in
  knowledge bases.
\newblock {\em arXiv preprint arXiv:1412.6575}.

\end{thebibliography}

\end{document}